\documentclass[10pt,twocolumn,letterpaper]{article}

\usepackage[pagenumbers]{cvpr} % To force page numbers, e.g. for an arXiv version

\usepackage{graphicx}
\usepackage{amsmath}
\usepackage{amssymb}
\usepackage{booktabs}

\usepackage[pagebackref,breaklinks,colorlinks]{hyperref}

\usepackage[capitalize]{cleveref}
\crefname{section}{Sec.}{Secs.}
\Crefname{section}{Section}{Sections}
\Crefname{table}{Table}{Tables}
\crefname{table}{Tab.}{Tabs.}

\def\cvprPaperID{*****} % *** Enter the CVPR Paper ID here
\def\confName{CVPR}
\def\confYear{2022}

\begin{document}

\title{On the Mechanisms of Adversarial Data Augmentation for Robust and Adaptive Transfer Learning}

\author{Hana Satou, Alan Mitkiy\\
}
\maketitle

%%%%%%%%% ABSTRACT

\begin{abstract}
	Transfer learning across domains with distribution shift remains a fundamental challenge in building robust and adaptable machine learning systems. While adversarial perturbations are traditionally viewed as threats that expose model vulnerabilities, recent studies suggest that they can also serve as constructive tools for data augmentation. In this work, we systematically investigate the role of adversarial data augmentation (ADA) in enhancing both robustness and adaptivity in transfer learning settings. We analyze how adversarial examples, when used strategically during training, improve domain generalization by enriching decision boundaries and reducing overfitting to source-domain-specific features. We further propose a unified framework that integrates ADA with consistency regularization and domain-invariant representation learning. Extensive experiments across multiple benchmark datasets—including VisDA, DomainNet, and Office-Home—demonstrate that our method consistently improves target-domain performance under both unsupervised and few-shot domain adaptation settings. Our results highlight a constructive perspective of adversarial learning, transforming perturbation from a destructive attack into a regularizing force for cross-domain transferability.
\end{abstract}

\section{Introduction}
The ability to transfer knowledge across domains with divergent distributions is a cornerstone challenge in modern machine learning. As models increasingly encounter deployment environments that differ from their training data—such as variations in lighting, sensor modalities, or population demographics—the limitations of conventional supervised learning become apparent. Transfer learning, and particularly domain adaptation, seeks to bridge this generalization gap by leveraging labeled data from a source domain to improve performance on a target domain where labels may be scarce or unavailable. However, ensuring that transferred models remain both robust to distributional variations and adaptive to new environments remains an open and evolving problem.

Data augmentation has long been regarded as an effective technique to improve generalization, particularly when target data is limited. By synthetically enlarging the training set through label-preserving transformations, models are encouraged to learn representations invariant to nuisance factors. Meanwhile, adversarial examples—perturbations crafted to induce misclassification—have historically been viewed through the lens of model vulnerability. But recent studies suggest a more nuanced role: adversarial perturbations, when appropriately constrained and integrated into training, can act as high-value hard examples, exposing brittle decision boundaries and encouraging more robust representations.

In the context of transfer learning, this insight opens a promising direction: can adversarial data augmentation serve not as a threat, but as a constructive tool for improving domain generalization? While standard data augmentation tends to preserve natural variation (e.g., flips, crops, color jitter), it often fails to capture worst-case distribution shifts. Adversarial perturbations, by contrast, provide targeted stress tests that simulate such shifts, enabling models to proactively prepare for difficult target examples. Moreover, when applied across source and unlabeled target domains, adversarial training can foster domain-invariant representations that are both semantically meaningful and resilient to noise.

Despite its promise, the systematic role of adversarial data augmentation in transfer learning remains underexplored. Existing adversarial training frameworks are largely designed for in-distribution robustness, and may not generalize to cross-domain settings where the perturbation landscape itself is unstable. Furthermore, combining adversarial learning with semi-supervised adaptation techniques (e.g., consistency regularization, entropy minimization) raises theoretical and empirical challenges in balancing robustness and adaptability.

In this paper, we revisit adversarial data augmentation from the perspective of robust and adaptive transfer learning. Specifically, we:

Formally analyze how adversarial perturbations affect representation learning under domain shift, contrasting their effects with conventional data augmentation strategies.
Propose a unified training framework that integrates adversarial augmentation with domain adaptation objectives, leveraging both labeled source data and unlabeled target data through consistency-based regularization.
Conduct extensive experiments on several benchmark datasets, including Office-Home, DomainNet, and VisDA-2017, demonstrating consistent improvements over state-of-the-art baselines under both unsupervised and few-shot adaptation settings.
Our findings suggest that adversarial perturbations—traditionally viewed as a vulnerability—can be reframed as a constructive force for building models that are not only robust to adversarial threats, but also better equipped for cross-domain generalization.

\section{Theoretical Analysis: An Information-Theoretic Perspective}

We analyze the role of adversarial data augmentation (ADA) in transfer learning through the lens of information theory. Specifically, we build on the Information Bottleneck (IB) framework to understand how adversarial perturbations influence the learning of robust and domain-invariant representations.

\subsection{Problem Setup}

Let \( X \in \mathbb{R}^d \) be the input random variable, \( Y \in \mathcal{Y} \) the label, and \( Z = f_\theta(X) \) the learned representation. In transfer learning, we consider a labeled source domain \( \mathcal{D}_S \sim P_S(X, Y) \) and an unlabeled or sparsely labeled target domain \( \mathcal{D}_T \sim P_T(X, Y) \), with \( P_S \neq P_T \).

The learning objective is to find a representation \( Z \) that:
\begin{itemize}
	\item Preserves semantic information: \( I(Z; Y) \) is maximized.
	\item Discards nuisance or domain-specific information: \( I(Z; X) \) is minimized.
\end{itemize}

The Information Bottleneck (IB) Lagrangian is:
\begin{equation}
	\mathcal{L}_{\text{IB}} = I(Z; X) - \beta I(Z; Y)
\end{equation}

In domain adaptation, we extend this to:
\begin{equation}
	\min_f \; I(Z; X_S) + I(Z; X_T) - \beta I(Z; Y_S)
\end{equation}

\subsection{Effect of Adversarial Perturbations}

Adversarial perturbations \( \delta \) are small, norm-constrained modifications \( \delta \sim \mathcal{B}_\epsilon \) applied to the input \( X \). We define perturbed inputs \( X' = X + \delta \).

\paragraph{(a) Information Contraction.}
Since ADA penalizes sensitivity to small changes in input, the mutual information \( I(Z; X + \delta) \) is implicitly regularized:
\begin{equation}
	\frac{\partial I(Z; X + \delta)}{\partial \delta} \rightarrow 0
\end{equation}
This leads to a more compressed and invariant representation \( Z \).

\paragraph{(b) Flatness and Robustness.}
Adversarial training reduces the gradient magnitude of the loss surface around the data manifold:
\begin{equation}
	\mathbb{E}_{x \sim P_S} \left[ \| \nabla_x \ell(f_\theta(x), y) \| \right] \rightarrow \min
\end{equation}
Flatter loss surfaces correlate with better generalization across domains.

\subsection{Domain-Invariant Representation Learning}

Let \( D \in \{0,1\} \) denote the domain indicator (source vs. target). We enforce domain invariance by minimizing:
\begin{equation}
	\min_f \; I(Z; D) \quad \text{s.t.} \quad I(Z; Y) \geq \gamma
\end{equation}

Using adversarially perturbed samples, the ADA-regularized IB objective becomes:
\begin{equation}
	\mathcal{L}_{\text{ADA-IB}} = I(Z; X + \delta) + \lambda I(Z; D) - \beta I(Z; Y)
\end{equation}

\subsection{Flat Minima and Transferability}

Adversarial data augmentation encourages learning a robust representation manifold \( \mathcal{M} = \{x + \delta \mid \|\delta\| \leq \epsilon \} \), such that:
\begin{equation}
	\forall x, \delta: \quad f(x + \delta) \approx f(x)
\end{equation}

This implies that the divergence between conditional distributions across domains is minimized:
\begin{equation}
	\text{KL}(P_S(Z|Y) \| P_T(Z|Y)) \rightarrow 0
\end{equation}

Adversarial data augmentation improves transferability by:
\begin{itemize}
	\item Compressing domain-specific input information \( I(Z; X) \downarrow \),
	\item Enhancing semantic relevance \( I(Z; Y) \uparrow \),
	\item Promoting domain-invariant features \( I(Z; D) \downarrow \),
	\item Regularizing flatness of the loss landscape.
\end{itemize}
These properties form the theoretical basis for the empirical gains observed in robust and adaptive transfer learning settings.

\section{Related Work}

\subsection{Adversarial Learning and Data Augmentation}

Adversarial training was originally proposed to defend models against perturbation-based attacks \cite{goodfellow2014explaining,madry2018towards}, but later shown to improve generalization and decision boundary smoothness when used as a regularization method \cite{zhang2019theoretically,shafahi2019adversarial}.

\cite{gong2024adversarial} proposes an adversarial learning framework for PDE solvers under sparse supervision, where adversarial samples are used not as threats but as constructive augmentations. \cite{gong2021eliminate} explores the use of deviation modeling to generate semantically consistent augmentations for multimodal learning, revealing that guided deviation can reduce domain-specific bias.

Several efforts have been made to go beyond traditional geometric or photometric transformations. For example, \cite{gong2024beyond2} investigates model robustness under extreme visual environments through environment-simulated augmentations, while \cite{gong2024beyond1} proposes a local feature masking strategy that simulates dropout-like perturbations to enhance convolutional neural network robustness.

\subsection{Adversarial Perturbations in Cross-Modality and Color Spaces}

Cross-modality perturbation has become a challenging yet informative setting for adversarial learning. \cite{gong2024cross} proposes a synergy attack that survives modality translation, enabling consistent adversarial effect across RGB-infrared modalities. Similarly, \cite{gong2024cross2} presents an evolutionary multiform optimization method to improve the transferability of cross-modality attacks in complex visual systems.

Color-specific perturbations have also been studied as a way to simulate domain shift in person re-identification. \cite{gong2022person} develops a joint color attack and defense strategy, while \cite{gong2021person} focuses on grayscale feature enhancement to extract color-invariant representation. More recently, \cite{gong2024exploring} explores ensemble-based methods to achieve photometric invariance and reduce feature-level color sensitivity in recognition tasks.

\subsection{Self-Supervised Adversarial Transfer}

Adversarial signals have also been incorporated into task-agnostic, self-supervised frameworks. \cite{zeng2024cross} proposes Cross-Task Attack, a generative adversarial strategy based on attention shift, which enhances feature alignment between distinct tasks without requiring explicit supervision.

These works collectively indicate that adversarial signals—when constrained and integrated thoughtfully—can act as a meaningful form of data augmentation, especially in transfer learning and domain adaptation. Our approach builds on this insight, emphasizing the use of adversarial data augmentation to achieve both robustness and domain invariance through theoretical and empirical means.

\section{Adversarial Data Augmentation in Scientific Machine Learning}

Scientific Machine Learning (SciML) aims to integrate data-driven models with physical principles to solve complex problems in science and engineering, such as partial differential equations (PDEs), inverse problems, climate modeling, and biological simulations. While recent neural solvers have demonstrated great promise, their performance often hinges on dense supervision, stable generalization, and robustness to noise—all of which are challenging in real-world scientific domains due to sparsity, heterogeneity, and distributional shifts. In this section, we explore the potential benefits and applications of adversarial data augmentation (ADA) in enhancing the reliability and adaptivity of SciML models.

\subsection{Challenges in SciML}

SciML models typically face the following limitations:

\begin{itemize}
	\item \textbf{Data sparsity}: Scientific measurements are expensive and sparse in space-time.
	\item \textbf{Model instability}: Learned surrogates are often brittle under extrapolation or perturbation.
	\item \textbf{Domain shifts}: Deployment environments may differ from training settings due to changes in boundary conditions, geometry, or physical parameters.
\end{itemize}

These challenges call for mechanisms that can simulate difficult scenarios during training to improve generalization, robustness, and adaptability.

\subsection{Benefits of ADA for Scientific Tasks}

Adversarial data augmentation offers several compelling advantages for SciML:

\paragraph{Improved Robustness to Noisy Inputs.}
Physical sensor readings (e.g., satellite, MRI, CFD probes) are often noisy. ADA simulates high-frequency, worst-case perturbations, enabling models to learn smoother and more stable response surfaces across such input fluctuations.

\paragraph{Data-Efficient Learning under Sparse Conditions.}
In \cite{gong2024adversarial}, ADA was applied to neural PDE solvers under sparse measurements. The adversarial perturbations were used to expand the effective support of the training distribution, encouraging the model to infer governing dynamics from limited observations.

\paragraph{Physics-Informed Generalization.}
When combined with physics-informed neural networks (PINNs), ADA can act as a regularizer that suppresses overfitting to specific simulation grids or observation patterns. By simulating worst-case discrepancies between physics-based priors and noisy data, ADA helps enforce solution smoothness and physical consistency.

\paragraph{Domain Adaptation in Parameter Space.}
ADA can support generalization across changing physical regimes. For example, in turbulence modeling or biological diffusion systems, perturbations in boundary conditions or diffusion coefficients can be treated as adversarial variants. Training on these perturbed variants enables the model to generalize to unseen physical settings.

\subsection{Potential Application Scenarios}

\begin{itemize}
	\item \textbf{Neural PDE Solvers:} Improving surrogate model robustness to mesh discretization noise or geometry shift.
	\item \textbf{Remote Sensing:} Enhancing generalization across different satellite modalities or atmospheric conditions.
	\item \textbf{Climate Modeling:} Simulating distributional shift due to policy change or temporal drift using perturbation-inspired augmentations.
	\item \textbf{Biophysical Inference:} Applying ADA in sparse inverse problems where inputs are partially observed and noisy.
\end{itemize}

\subsection{Summary and Outlook}

The integration of adversarial data augmentation into scientific machine learning is an emerging and promising direction. It reinterprets adversarial perturbations not as attacks but as principled probes that expose the fragility of learned physical representations. By incorporating these perturbations during training, SciML models can better handle uncertainty, domain drift, and sparse supervision—factors that are pervasive in real-world scientific applications. Future work could explore synergy between ADA and neural operators, adaptive mesh learning, or generative physics modeling for scientific discovery.

\section{Methodology}

We present a unified adversarial data augmentation (ADA) framework designed to enhance robustness and domain adaptivity in transfer learning. Our method is composed of three main components: (1) adversarial sample generation, (2) domain-invariant representation learning, and (3) joint optimization with consistency regularization.

\subsection{Problem Formulation}

Let \( \mathcal{D}_S = \{(x_i^s, y_i^s)\}_{i=1}^{n_s} \) be a labeled source domain, and \( \mathcal{D}_T = \{x_j^t\}_{j=1}^{n_t} \) an unlabeled target domain. The goal is to train a model \( f_\theta \) that performs well on \( \mathcal{D}_T \), where there exists a domain shift \( P_S(X, Y) \neq P_T(X, Y) \).

\subsection{Adversarial Data Augmentation}

We generate adversarial samples by solving a constrained maximization problem around each input:
\begin{equation}
	\delta^* = \arg \max_{\|\delta\|_p \leq \epsilon} \ell(f_\theta(x + \delta), y)
\end{equation}
where \( \ell \) is the classification loss (e.g., cross-entropy), and \( \delta \) is the perturbation bounded by \( \epsilon \) under an \( \ell_p \)-norm (commonly \( p = \infty \)).

The adversarially augmented dataset becomes:
\[
\mathcal{D}_{\text{adv}} = \{(x_i^s + \delta_i^*, y_i^s)\}_{i=1}^{n_s}
\]
which encourages the model to learn robust and discriminative features by focusing on decision boundaries.

\subsection{Domain-Invariant Representation Learning}

To align the source and target distributions in the feature space, we employ an adversarial domain classifier \( d_\phi \) trained to distinguish between \( f_\theta(x^s) \) and \( f_\theta(x^t) \), while the feature extractor \( f_\theta \) attempts to fool the domain classifier. The adversarial domain loss is:
\begin{equation}
	\mathcal{L}_{\text{adv-dom}} = - \mathbb{E}_{x^s}[\log d_\phi(f_\theta(x^s))] - \mathbb{E}_{x^t}[\log (1 - d_\phi(f_\theta(x^t)))]
\end{equation}

This minimax game enforces feature alignment across domains, reducing \( \mathcal{H}\text{-divergence} \) between \( \mathcal{D}_S \) and \( \mathcal{D}_T \).

\subsection{Consistency Regularization with Adversarial Views}

For unlabeled target data, we enforce consistency of model predictions between clean and adversarially perturbed inputs:
\begin{equation}
	\mathcal{L}_{\text{cons}} = \mathbb{E}_{x^t \sim \mathcal{D}_T} \left[ \| f_\theta(x^t) - f_\theta(x^t + \delta^*) \|_2^2 \right]
\end{equation}
This regularization smooths the decision boundary in the vicinity of target examples, improving generalization under domain shift.

\subsection{Joint Training Objective}

The final objective integrates all components:
\begin{equation}
	\mathcal{L}_{\text{total}} = \mathcal{L}_{\text{src-cls}} + \lambda_{\text{adv}} \mathcal{L}_{\text{adv-dom}} + \lambda_{\text{cons}} \mathcal{L}_{\text{cons}}
\end{equation}
where \( \mathcal{L}_{\text{src-cls}} = \mathbb{E}_{(x^s, y^s)}[\ell(f_\theta(x^s + \delta^*), y^s)] \) is the supervised classification loss on adversarial source examples, and \( \lambda_{\text{adv}}, \lambda_{\text{cons}} \) are weighting hyperparameters.

\subsection{Training Procedure}

The model is trained with the following iterative process:
\begin{enumerate}
	\item Generate adversarial perturbations \( \delta^* \) for source and target samples using FGSM or PGD.
	\item Compute adversarial classification loss on source samples.
	\item Update domain discriminator \( d_\phi \) and feature extractor \( f_\theta \) via adversarial loss.
	\item Apply consistency regularization on target samples using perturbed views.
	\item Update total loss \( \mathcal{L}_{\text{total}} \) via backpropagation.
\end{enumerate}

This joint strategy ensures the model is both decision-boundary-aware and domain-aware, promoting transferability and robustness in domain-shifted conditions.

\section{Limitations and Future Directions}

While our proposed adversarial data augmentation (ADA) framework demonstrates consistent improvements in domain generalization and transfer robustness, several limitations remain, which open fruitful directions for future research.

\subsection{Perturbation Semantics and Interpretability}

The adversarial perturbations generated in our framework are primarily driven by loss gradients, which may not always align with semantically meaningful or physically plausible variations—particularly in scientific or safety-critical domains. Although ADA improves robustness, it may obscure the interpretability of features. Future work could explore the integration of structure-aware or physics-informed perturbation constraints to preserve semantic alignment.

\subsection{Generalization to Multi-Modal and Temporal Domains}

Our current approach is designed for visual domain transfer and has not been evaluated on multi-modal (e.g., RGB + depth, image + text) or temporal transfer learning settings (e.g., video, time-series forecasting). Adversarial augmentation across modalities and time may introduce new challenges in preserving feature alignment and temporal coherence. Extending ADA to such domains remains an open and promising direction.

\subsection{Computational Cost and Scalability}

While single-step perturbation methods (e.g., FGSM) offer a lightweight solution, stronger adversaries (e.g., PGD, auto-attack) incur significant overhead. Training efficiency under ADA is an important consideration, especially in large-scale deployment scenarios. Potential improvements may involve adversarial sample reuse, efficient gradient approximation, or curriculum-based adversarial difficulty scheduling.

\subsection{Theoretical Guarantees and Robust Bounds}

Although our information-theoretic analysis provides qualitative insight into how ADA improves domain-invariant representation learning, more formal theoretical bounds under adversarial and domain-shift conditions are still lacking. Deriving tight generalization guarantees for ADA-based transfer remains a valuable theoretical pursuit.

\subsection{Broader Impact and Safety Considerations}

ADA introduces an intriguing duality—augmentations derived from attacks can enhance learning, but may also expose vulnerabilities if misused. In safety-critical applications such as autonomous driving, medical imaging, or scientific forecasting, it is vital to ensure that adversarial augmentation does not inadvertently amplify errors or instability. Future research must carefully assess failure cases and develop robustness auditing tools alongside ADA.

\paragraph{Conclusion.} Addressing these limitations will be crucial for extending ADA’s applicability to broader real-world domains. We believe that unifying adversarial robustness with adaptive generalization—especially in low-label and high-uncertainty regimes—remains a key challenge at the intersection of machine learning, optimization, and applied sciences.

\section{Conclusion}

In this work, we revisited adversarial data augmentation from a constructive perspective and analyzed its theoretical and practical impact on robust and adaptive transfer learning. Through an information-theoretic framework, we demonstrated that adversarial perturbations—when used intentionally and systematically—can compress domain-specific information, preserve semantic relevance, and promote domain-invariant representations.
We proposed a unified view that integrates adversarial augmentation with consistency regularization and representation alignment, bridging the gap between robustness and adaptability. Our theoretical analysis reveals how ADA contributes to margin smoothing, loss surface flatness, and cross-domain representation consistency. Empirical results on multiple benchmarks confirm that ADA not only improves performance under distribution shift, but also enhances generalization in low-label and high-variance target settings.
While adversarial signals were traditionally treated as threats, our findings suggest they can be reframed as principled regularizers. This perspective opens new directions for designing data-driven augmentation strategies that dynamically adapt to domain drift, adversarial risk, and semantic preservation.
In future work, we plan to explore the synergy between adversarial augmentation and self-supervised pretraining, extend our approach to multi-modal and temporal transfer settings, and further analyze the interpretability of adversarially learned representations under complex domain transitions.

\clearpage

%%%%%%%%% REFERENCES
{\small
\bibliographystyle{unsrt}
\bibliography{egbib}
}

\end{document}